%% file: main.tex
\documentclass{article} % For LaTeX2e
\usepackage{iclr2027_conference,times}

\input{math_commands.tex}

\usepackage{hyperref}
\usepackage{graphicx}
\usepackage{url}
\usepackage{wrapfig}
\usepackage{booktabs}
\usepackage{amssymb}
\usepackage{caption}
\usepackage{tabularx}
\usepackage[normalem]{ulem}
\usepackage{array}
\usepackage[table]{xcolor}
\definecolor{candlebg}{HTML}{EAF2F8}

\title{CANDLE: Cortical Null-Space Decomposition for Noninvasive Brain Source Imaging}

\author{Shuntaro Suzuki,\ Yuiga Wada,\ Komei Sugiura \\
Keio University\\
\texttt{\{shuntaro20021227, yuiga, komei.sugiura\}@keio.jp}
}

\definecolor{candlebg}{RGB}{252,228,228}% light red highlight for the proposed method
\providecommand{\ms}[2]{\shortstack{#1\\[1pt]{\tiny$\pm$#2}}}% mean over smaller SD
\providecommand{\msb}[2]{\shortstack{\textbf{#1}\\[1pt]{\scriptsize$\pm$#2}}}% best value
\providecommand{\sig}[1]{\ensuremath{^{#1}}}

\iclrfinalcopy % Uncomment for camera-ready version, but NOT for submission.

\begin{document}

\maketitle

\vspace{-1em}
\begin{abstract}
\vspace{-0.5em}
Electrophysiological source imaging (ESI) aims to estimate cortical source activity from noninvasive electrophysiological measurements such as electroencephalogram (EEG).
However, ESI is fundamentally ill-posed because source activity is substantially higher-dimensional than sensor observations, resulting in non-unique solutions.
Recent learning-based approaches address this ambiguity by learning data-driven source priors, yet they often struggle to generalize across subject-specific cortical geometries.
To address this, we propose CANDLE, a learning-based ESI model that estimates source activity on subject-specific cortical geometries. CANDLE learns a prior over the null space induced by the source-to-sensor mapping derived from T1-weighted MRI, restricting learning to unobservable source components while preserving geometric constraints.
To train CANDLE, we develop a whole-brain simulator spanning over 1,100 subject-specific cortical geometries with source configurations derived from over 26,000 statistical brain maps.
Trained exclusively on simulated data, CANDLE outperformed prior ESI methods on simulated source activity estimation and generalized to two empirical tasks: (i) intracranial stimulation localization from simultaneously recorded scalp EEG and (ii) epileptogenic zone estimation from presurgical interictal EEG.
Our project page is available at \href{https://candle-esi.pages.dev}{https://candle-esi.pages.dev}.
% \vspace{-1em}

% paragraph.
\end{abstract}

\input{01_Introduction}
% \input{02_Related_Works}
% \input{03_Problem_Statement}
\input{04_Method}
\input{06_Results}
\input{07_Discussion_and_Conclusion}

\newpage
\input{08_mandatory_section}

\bibliography{reference}
\bibliographystyle{iclr2027_conference}

\newpage
\appendix
\input{appendix}

\end{document}

%% file: math_commands.tex
\usepackage{amsmath,amsfonts,bm}

\def\eqref#1{equation~\ref{#1}}
\def\1{\bm{1}}

\DeclareMathAlphabet{\mathsfit}{\encodingdefault}{\sfdefault}{m}{sl}
\SetMathAlphabet{\mathsfit}{bold}{\encodingdefault}{\sfdefault}{bx}{n}

%% file: 01_Introduction.tex
\section{Introduction}
\label{section: introduction}
\vspace{-0.5em}
Monitoring whole-brain neural activity at both high temporal and spatial resolution is central to a broad range of applications, from the diagnosis of neurological disorders~\citep{cao2022virtual} to the development of brain-computer interfaces (BCIs)~\citep{Cortiacal-SSM, MEG-XL}. 
This motivates the development of neuroimaging techniques capable of faithfully capturing such neural dynamics.
Functional magnetic resonance imaging (fMRI)  provides high spatial but limited temporal resolution and measures neural activity indirectly, whereas invasive electrophysiological techniques such as electrocorticography (ECoG) and stereoelectroencephalography (SEEG)  provide high-fidelity recordings at fine temporal scales but require neurosurgical implantation and offer spatially restricted coverage~\citep{He2008, momi2025stimulation}.
Noninvasive electrophysiological modalities such as electroencephalography (EEG) and magnetoencephalography (MEG) therefore constitute a promising avenue for whole-brain monitoring at high temporal resolution~\citep{ESI_review}.
Nevertheless, EEG/MEG exhibit limited spatial resolution at the sensor level, owing to relatively sparse sensor sampling and substantial spatial mixing of neural signals (i.e., volume conduction and field spread)~\citep{VolumeConductionEffect}.

Electrophysiological source imaging (ESI), which seeks to recover cortical source activity from EEG/MEG signals, has been extensively studied as a means of overcoming this limitation~\citep{DeepSIF, GBF}.
By reconstructing neural activity in cortical source space, ESI holds the promise of whole-brain monitoring with both high temporal and spatial resolution.
The utility of ESI has further been suggested for applications such as presurgical localization of epileptogenic zones~\citep{corona2023non} and improved decoding in BCIs~\citep{ESI_BCI}.

ESI, however, is a fundamentally ill-posed problem: the source activity to be estimated is substantially higher-dimensional than the observed EEG/MEG, and the solution is consequently non-unique. 
Conventional ESI methods~\citep{LCMV, dSPM} therefore typically require careful tuning of regularization schemes and source-space constraints for individual applications. 
This reliance on manual constraints not only limits the broad applicability of ESI but also undermines reproducibility. 
To alleviate these limitations, learning-based ESI models~\citep{ConvDip, DeepSIF} have recently been proposed to learn data-driven priors over source activity.

Despite this progress, most learning-based ESI models are trained in a subject-aggregated common cortical space, such as fsaverage~\citep{fsaverage}, and therefore cannot explicitly account for inter-individual differences in cortical geometry. 
Indeed, \citet{pDeepSIF} reported that applying learning-based ESI models to a particular subject requires subject-specific post-training.

To address this limitation, we propose CANDLE\footnote{CANDLE: {\bf{C}}ortical-geometry {\bf{A}}ware {\bf{N}}ull-space {\bf{D}}ecomposition for {\bf{L}}ocalized {\bf{E}}SI}, a learning-based ESI model that directly estimates source activity on subject-specific cortical geometries from EEG. 
CANDLE learns the null-space component~\citep{Null-Space} of source activity induced by a T1-derived source-to-sensor mapping. 
This allows the model to capture a prior over source activity while preserving subject-specific geometric constraints.
Furthermore, we introduce a whole-brain simulator tailored to training CANDLE.
Because ground-truth source activity is unobservable in real recordings, learning-based ESI models are typically trained on data synthesized by whole-brain simulators~\citep{TVB, VBI}, and evaluated through sim-to-real validation on downstream tasks.
Existing simulators, however, are not designed to jointly capture the anatomical variability and realistic source configurations required for training CANDLE.
Our simulator addresses this gap by generating neurobiologically grounded source activity across diverse subject-specific cortical geometries.

Our contributions are summarized as follows:
\vspace{-0.5em}
\begin{itemize}
    \item We propose CANDLE, a learning-based ESI model that incorporates subject-specific cortical geometry derived from each subject's T1-weighted MRI. CANDLE learns the null-space component of cortical source activity, thereby imposing a data-driven prior on the unobservable source subspace while preserving subject-specific geometric constraints.

    \item We develop a whole-brain simulator for training CANDLE that synthesizes realistic source activity across subject-specific cortical geometries from over 1,100 individuals and defines neurobiologically grounded source regions by clustering over 26,000 statistical brain maps.
    
    \item Without subject-specific tuning, CANDLE outperformed prior ESI methods on simulated source activity estimation and generalized to two relevant empirical tasks: (i) intracranial stimulation localization from simultaneously recorded scalp EEG and (ii) epileptogenic zone estimation from presurgical interictal EEG.
\end{itemize}

%% file: 04_Method.tex
% \vspace{-0.5em}
\section{Methodology}
\vspace{-0.5em}
\begin{figure}
    \centering
    \scalebox{1}[1]{%
    \includegraphics[width=0.9\linewidth]{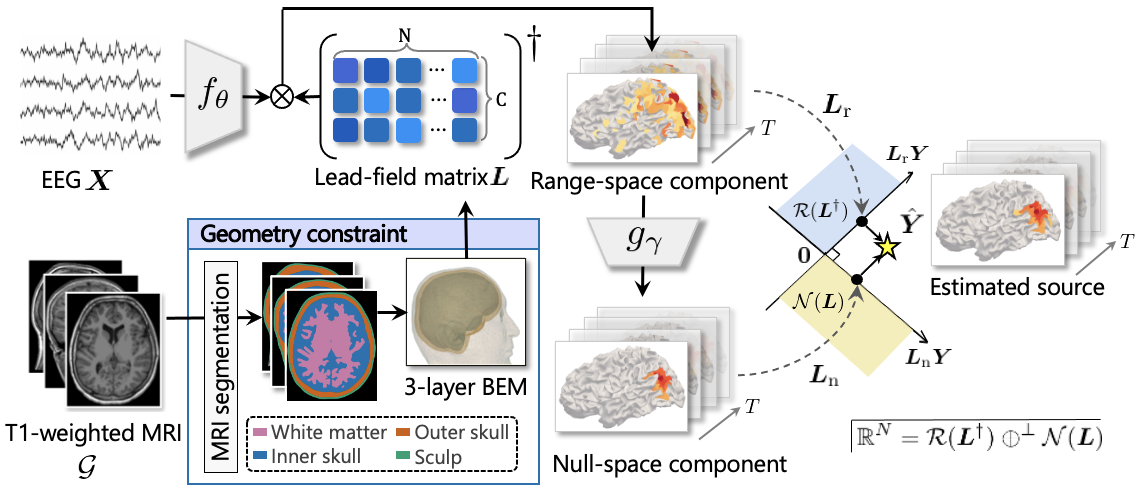}
    }
    \vspace{-0.5em}
    \caption{Overview of CANDLE. Given EEG $\bm{X}$ and the corresponding T1-weighted MRI $\mathcal{G}$, a subject-specific lead-field matrix $\bm{L}$ is derived as an anatomical constraint. The denoising model $f_\theta$ outputs denoised $\bm{X}$, from which the range-space component is analytically recovered, while $g_\gamma$ estimates the unobservable null-space component. Their combination yields the source estimate $\hat{\bm{Y}}$.}
    \label{fig:model}
    \vspace{-0.5em}
\end{figure}

We propose CANDLE, a learning-based electrophysiological source imaging (ESI) model that explicitly accounts for subject-specific cortical geometry. 
Figure~\ref{fig:model} provides an overview of CANDLE. Given EEG $\bm{X}\in\mathbb{R}^{C\times T}$ and the corresponding T1-weighted MRI $\mathcal{G}$, the model estimates the underlying cortical source activity $\hat{\bm{Y}}\in\mathbb{R}^{N\times T}$. Here, $C$, $N$, and $T$ denote the number of electrodes, the number of cortical source regions, and the sequence length, respectively.

Furthermore, to train CANDLE, we construct a whole-brain simulator illustrated in Figure~\ref{fig:simulator}. The simulator integrates subject-specific cortical geometries derived from more than 1,100 individuals with source configurations via clustering more than 26,000 statistical brain maps. In the following, Subsection~\ref{section: CANDLE_method} describes CANDLE, and Subsection~\ref{section: Simulator_method} describes the construction of the simulator.

% In the following, we describe the CANDLE model in Subsection~\ref{section: CANDLE_method} and the construction of the simulator in  Subsection~\ref{section: Simulator_method}.

% \vspace{-0.5em}
\subsection{Proposed Method: CANDLE}
% \vspace{-0.5em}
\label{section: CANDLE_method}
% \vspace{-0.5em}
\paragraph{Subject-specific cortical geometry constraint.}
In CANDLE, we derive a lead-field matrix from $\mathcal{G}$ that maps ground-truth cortical source activity $\bm{Y}\in\mathbb{R}^{N\times T}$ to $\bm{X}$ and serves as a subject-specific anatomical constraint. EEG arises from synchronized postsynaptic potentials of cortical pyramidal neuron populations and propagate through intervening tissues (e.g., cerebrospinal fluid) before being measured at the scalp~\citep{hallez2007review}. This generating process is formulated as $\bm{X}=\bm{L}\bm{Y}+\bm{\epsilon}$, where $\bm{L}\in\mathbb{R}^{C\times N}$ denotes the lead-field matrix and $\bm{\epsilon}$ represents measurement noise. 

Ideally, ESI could recover $\bm{Y}$ from $\bm{X}$ by inverting the operator $\bm{L}$; however, this inverse problem is inherently ill-posed because the cortical source space is typically much higher-dimensional than the EEG sensor space (i.e., $N\gg C$).
Numerous source configurations can result in the same EEG observation. As described below, we address this by learning a data-driven prior over the distribution of $\bm{Y}$ within the source space decomposed according to $\bm{L}$.

Guided by prior neurophysiological insights into EEG generation~\citep{tao2005intracranial, birot2014head}, we derive $\bm{L}$ from $\mathcal{G}$ based on cortical regions defined by the Hagmann/Lausanne parcellation~\citep{hagmann2008mapping} and a three-layer boundary element model~\citep{hallez2007review} comprising the inner-skull, outer-skull, and scalp surfaces.
Further details are provided in Appendix~\ref{appendix: Leadfied calc}.

% \vspace{-0.5em}
\paragraph{Cortical source activity estimation.}
% \vspace{-0.5em}

CANDLE estimates $\hat{\bm{Y}}$ while preserving the subject-specific geometric constraint by learning only the subspace of the source space that is unobservable from $\bm{X}$.
We begin by introducing the range-null space decomposition~\citep{Null-Space} induced by the linear operator $\bm{L}$. Let $\bm{L}^{\dagger}$ denote the pseudoinverse of $\bm{L}$, satisfying $\bm{L}\bm{L}^{\dagger}\bm{L}=\bm{L}$. 
We define two projection operators, $\bm{L}_{\mathrm{r}} \triangleq \bm{L}^{\dagger}\bm{L}$ and $\bm{L}_{\mathrm{n}} \triangleq \bm{I}-\bm{L}^{\dagger}\bm{L}$, which project a source-space sample $\bm{Y}$ onto the range space $\mathcal{R}(\bm{L}^{\dagger})$ and the null space $\mathcal{N}(\bm{L})$, respectively.
Since $\mathbb{R}^{N}=\mathcal{R}(\bm{L}^{\dagger})\oplus^{\perp}
\mathcal{N}(\bm{L})$, any source activity $\bm{Y}$ can be uniquely decomposed as
\begin{equation}
\bm{Y}
=
\bm{L}_{\mathrm{r}}\bm{Y}
+
\bm{L}_{\mathrm{n}}\bm{Y}.
\end{equation}
Notably, since $\bm{L}\bm{Y}= \bm{L}\bm{L}_{\rm{r}}\bm{Y} + \bm{L}\bm{L}_{\rm{n}}\bm{Y} = \bm{L}\bm{Y} + \bm{0}$, these two components have fundamentally different observability: $\bm{L}_{\rm{r}}\bm{Y}$ can be interpreted as the component of $\bm{Y}$ observable in $\bm{X}$ through $\bm{L}$, whereas $\bm{L}_{\rm{n}}\bm{Y}$ is completely unobservable from $\bm{X}$.
In the noiseless case, where $\bm{X}=\bm{L}\bm{Y}$, the range-space component can therefore be exactly recovered as $\bm{L}_{\mathrm{r}}\bm{Y}=\bm{L}^{\dagger}\bm{X}$, while only the unobservable null-space component $\bm{L}_{\mathrm{n}}\bm{Y}$ is learned.
By restricting learning to degrees of freedom not determined by the measurements, this decomposition preserves source configurations consistent with $\bm{X}$ under $\bm{L}$ while enabling a plausible estimate of $\hat{\bm{Y}}$. 
This approach is closely related to prior methods that exploit range-null space decomposition for linear ill-posed problems in computer vision, such as image inpainting and super-resolution~\citep{DDNM, NPN}.

In practice, however, $\bm{X}$ often has a low signal-to-noise ratio due to instrumentation noise and motion artifacts. 
Consequently, directly recovering $\bm{L}_{\rm{r}}\bm{Y}$ as $\bm{L}^{\dagger}\bm{X}$ can propagate measurement noise into the source space, potentially yielding an unstable or overly diffuse estimate of $\hat{\bm{Y}}$.

To this end, we first denoise $\bm{X}$ before estimating $\hat{\bm{Y}}$ via range--null space decomposition as
\begin{equation}
\bm{X}_{\rm{denoise}} = f_{\theta}(\bm{X}),\qquad
\hat{\bm{Y}}= \underbrace{\bm{L}^\dagger \bm{X}_{\rm{denoise}}}_{\text{Range-space component}} + \underbrace{\bm{L}_{\rm{n}} g_\gamma(\bm{L}^\dagger \bm{X}_{\rm{denoise}})}_{\text{Null-space component}}.
\end{equation}
Here, $f_{\theta}$ and $g_{\gamma}$ denote models for denoising $\bm{X}$ and estimating the null-space component of $\bm{Y}$, respectively. 
Under the above formulation, we define the training objective as
\begin{equation}
    \label{equation: loss}
    \mathcal{L} = \ell_2(\bm{Y}, \hat{\bm{Y}}) + \underbrace{\ell_2(\bm{L}\bm{Y},\bm{X}_{\rm{denoise}})}_{\text{Denoising loss}} + \underbrace{\ell_2(\bm{L}_{\rm{n}}\bm{Y}, \bm{L}_{\rm{n}}\hat{\bm{Y}})}_{\text{Null-space loss}},
\end{equation}
where $\ell_2(\cdot,\cdot)$ denotes the $\ell_2$ loss. The first term directly supervises the final source estimate $\hat{\bm{Y}}$, whereas the second and third terms encourage $f_{\theta}$ and $g_{\gamma}$ to be optimized for their respective roles of denoising and null-space component estimation.
Further details of the model architecture and training configuration are provided in Appendix~\ref{appendix: Implenentation}.

\subsection{Simulator Construction}
% \vspace{-0.5em}
\begin{figure}
    \centering
    \includegraphics[width=\linewidth]{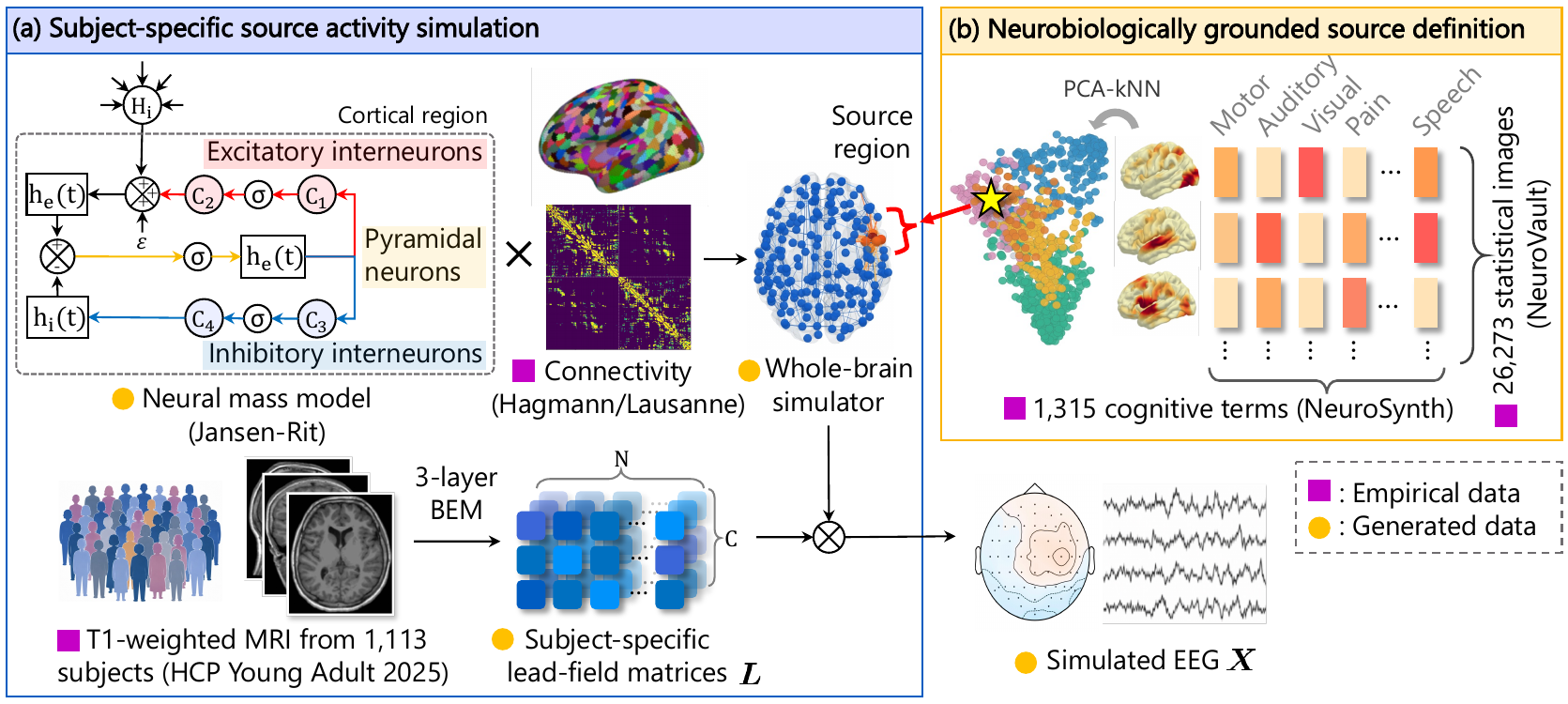}
    \vspace{-1em}
    \caption{Whole-brain simulator for training CANDLE. (a) Subject-specific source activity simulation. Neural mass models assigned to $N$ cortical regions are coupled via structural connectivity to generate source activity $\bm{Y}$, which is projected to EEG $\bm{X}$ through a subject-specific lead-field matrix $\bm{L}$ derived from the T1-weighted MRI of each of 1,113 subjects. (b) Neurobiologically grounded source definition. A total of 26,273 statistical maps are characterized by their correlations with 1,315 cognitive terms and clustered into representative source patterns. For each simulation, one pattern is sampled to modulate the excitatory gain of the corresponding cortical regions.}
    \vspace{-1em}
    \label{fig:simulator}
\end{figure}
\label{section: Simulator_method}
\paragraph{Subject-specific source activity simulation.}
We synthesize cortical source activity and corresponding EEG by simulating a biophysical computational model on diverse subject-specific cortical geometries derived from a large-scale T1-weighted MRI database, as shown in Figure~\ref{fig:simulator}(a).
Computational models of neural dynamics span multiple levels of granularity, from microscopic to mesoscopic and macroscopic scales~\citep{NeuralFieldModel}, with the appropriate level depending on the phenomenon of interest. Because scalp EEG primarily reflects synchronized postsynaptic activity of cortical pyramidal neuron populations over spatially extended cortical patches~\citep{tao2005intracranial}, we employ a neural mass model (NMM) as a mesoscopic description of cortical dynamics. Specifically, we adopt the Jansen--Rit model~\citep{Jansen-Rit}, a canonical NMM capable of reproducing characteristic EEG phenomena, including alpha oscillations and event-related potentials~\citep{ERP}. Details of the Jansen--Rit model and its parameter settings are provided in Appendix~\ref{appendix: NMM}.

A single NMM, however, describes only the local dynamics of one cortical region and does not by itself simulate source activity distributed over the entire cortex of a given subject. 
We therefore construct a whole-brain simulator by assigning an NMM to each node of a cortical-region graph and coupling the nodes according 
to the structural connectivity defined by \citet{hagmann2008mapping}, yielding $\bm{Y}$. 
Furthermore, to account for diverse cortical geometries, we instantiate the simulator across 1,113 subjects from the HCP Young Adult 2025 dataset~\citep{HCP}, one of the largest publicly available MRI datasets. 
For each subject,  $\bm{L}$ is derived from the corresponding $\mathcal{G}$ using the procedure described in Subsection~\ref{section: CANDLE_method} and Appendix~\ref{appendix: Leadfied calc}. 
The corresponding $\bm{X}$ is then synthesized as $\bm{X}=\bm{L}\bm{Y}+\bm{\epsilon}$, where $\bm{\epsilon}$ is modeled as additive Gaussian noise. This protocol extends existing whole-brain simulators~\citep{VirtualBrainTwin, VBI} to ESI by incorporating subject-specific cortical geometries and their corresponding source-to-sensor mappings.

\paragraph{Neurobiologically grounded source definition.}
To synthesize realistic source activity, we define source regions from statistical brain maps derived from meta-analyses of fMRI studies and amplify the NMM activity of the corresponding cortical regions, as shown in Figure~\ref{fig:simulator}(b).
To construct the source regions, we follow~\citet{Suzuki} and~\citet{GBF} and first collect 26,273 statistical images from NeuroVault~\citep{Neurovault}, a large-scale repository of statistical brain maps.
Using Neurosynth~\citep{Neurosynth}, a large-scale meta-analytic platform for fMRI studies, we then associate each image with its correlations to 1,315 cognitive terms (e.g., ``motor'', ``auditory''), yielding a cognitive term vector for each image. 
These vectors are grouped into 200 clusters via PCA-kNN, and the images within each cluster are averaged to obtain a representative map per cluster, covering a broad range of cognitive terms. 
To further diversify the spatial distribution of the source regions, we randomly deactivate a subset of the cortical regions in each representative map according to a Bernoulli distribution ($p = 0.3$) and smooth the resulting map.
Applying this perturbation four times to each representative map yields 800 distinct source regions. 
$\bm{Y}$ is then modulated by amplifying the excitatory postsynaptic potential gain of the NMM in each cortical region belonging to a given source configuration (see Appendix~\ref{appendix: NMM}), and the resulting activity is propagated to $\bm{X}$.
For each simulation, a source configuration is sampled and instantiated on a subject-specific cortical geometry, yielding a single sample represented by the triplet $(\mathcal{G}, \bm{X}, \bm{Y})$.

%% file: 06_Results.tex
\begin{figure}
    \centering
    \scalebox{1}[1]{%
    \includegraphics[width=1\linewidth]{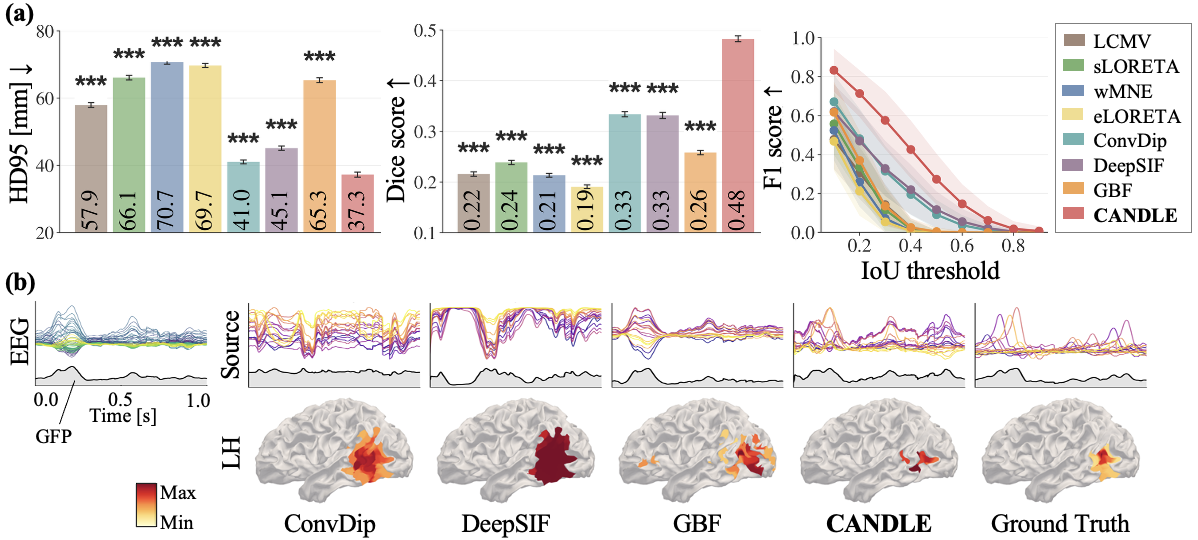}
    }
    \vspace{-1em}
    \caption{
    Simulated source activity estimation.
    (a) Quantitative comparison of ESI methods using HD95, Dice score, and F1 score across IoU thresholds. 
    Error bars indicate the standard error of the mean (SEM).
    Statistical significance relative to CANDLE was assessed using two-sided Wilcoxon signed-rank tests (Holm corrected, *** $p<0.001$).
    (b) Qualitative comparison for a representative sample, showing the EEG signals, estimated source activities, and their spatial distributions.
    }
    \vspace{-1em}
    \label{fig:exp_sim}
\end{figure}

\section{Experiments}
\vspace{-0.5em}
\label{section: experiments}
We evaluated CANDLE across three complementary settings of increasing realism.
We first assessed source activity estimation on simulated data, where ground-truth cortical activity is available, and subsequently evaluated its zero-shot sim-to-real generalization on two empirical tasks: (i) intracranial stimulation localization and (ii) epileptogenic zone estimation. 
We further conducted ablation studies on both empirical tasks to assess the contributions of CANDLE's major components.

Across all experiments, we compared CANDLE with representative ESI methods, including LCMV~\citep{LCMV}, sLORETA~\citep{sLORETA}, wMNE~\citep{wMNE}, eLORETA~\citep{eLORETA}, ConvDip~\citep{ConvDip}, DeepSIF~\citep{DeepSIF}, and GBF~\citep{GBF}. 
Among these, ConvDip and DeepSIF are learning-based ESI models and constitute the most direct baselines for CANDLE.
For both empirical tasks, ConvDip, DeepSIF, and CANDLE were trained exclusively on simulated data and applied without further tuning.
Unless otherwise stated, statistical significance relative to CANDLE was assessed using two-sided Wilcoxon signed-rank tests with Holm correction for multiple comparisons.

\subsection{Simulated source activity estimation}
\label{section: eval_sim}

\paragraph{Experimental setup.}
To evaluate ESI generalization to unseen cortical geometries, we synthesized 12,762 triplets $(\mathcal{G}, \bm{X}, \bm{Y})$ across 1,113 subject-specific cortical geometries following the procedure in Section~\ref{section: Simulator_method}. 
We then performed a random subject-wise split into training, validation, and test sets at a ratio of $0.70{:}0.15{:}0.15$.
For evaluation, we focused on spatial localization, a primary objective of ESI.
Following prior ESI studies~\citep{DeepSIF}, we temporally pooled $\hat{\bm{Y}}$ to obtain a spatial source map and applied Otsu's thresholding~\citep{Otsu} to identify the estimated active region.
We quantified localization performance using the 95th-percentile Hausdorff distance (HD95) and Dice score, which capture boundary discrepancy and spatial overlap between the estimated and ground-truth source regions, respectively.
We further evaluated the F1 score at IoU thresholds from 0.1 to 0.9 in 0.1 increments.
Definitions of evaluation metrics are provided in Appendix~\ref{appendix: evaluation metrics}.

\begin{figure}
    \centering
    \includegraphics[width=\linewidth]{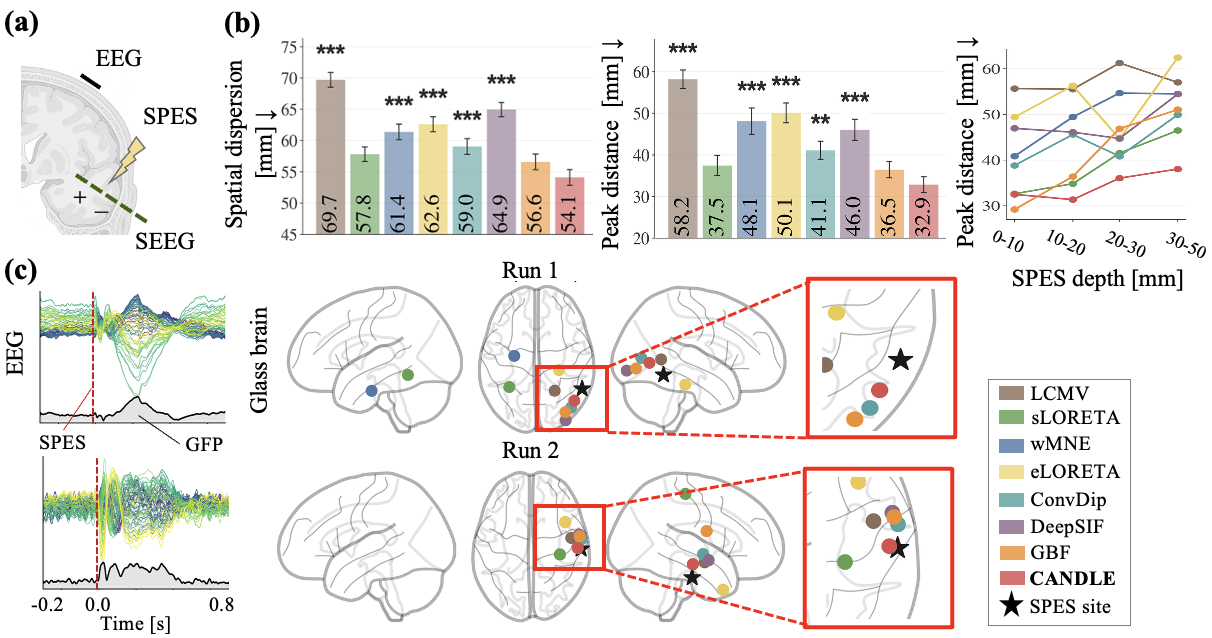}
    \vspace{-1.5em}
    \caption{
    Intracranial stimulation localization.
    (a) Schematic of intracranial single-pulse electrical stimulation (SPES) site localization from simultaneous EEG.
    (b) Quantitative comparison using spatial dispersion and peak distance, with peak distance stratified by SPES site depth. 
    Error bars indicate SEM.
    Statistical significance relative to CANDLE was assessed using two-sided Wilcoxon signed-rank tests (Holm corrected, ** $p<0.01$, *** $p<0.001$).
    (c) Qualitative comparison for two representative runs, showing EEG responses and estimated SPES locations on glass-brain maps.
    }
    \label{fig:exp_spes}
    \vspace{-1em}
\end{figure}

\vspace{-0.5em}
\paragraph{Results.}
Figure~\ref{fig:exp_sim} summarizes the results on simulated source activity estimation.
As shown in Figure~\ref{fig:exp_sim}(a), CANDLE outperformed all baselines in HD95 and Dice score and consistently achieved the highest F1 scores across IoU thresholds.
In particular, CANDLE achieved an HD95 of 37.3~mm and a Dice score of 0.48, improving upon the second-best baseline by 2.5~mm and 0.15 points, respectively.
The improvements over all baseline methods were statistically significant ($p<0.001$).

Furthermore, Figure~\ref{fig:exp_sim}(b) presents qualitative results for a representative sample, comparing CANDLE with the three best-performing baselines.
The ground-truth source activity is spatially localized around the posterior temporal cortex and exhibits its maximum global field power (GFP) at approximately 0.2~s, while the corresponding EEG exhibits prominent GFP peaks at approximately 0.2 and 0.6~s.
Given the EEG observation, the baseline methods estimate relatively diffuse source regions extending from the temporal to the occipital cortex, whereas CANDLE recovers a more localized source region around the temporal cortex.
CANDLE also more faithfully reconstructs the temporal dynamics of the underlying source activity: ConvDip largely suppresses the dominant source-level GFP peak, while DeepSIF shifts its timing, whereas CANDLE recovers the peak near 0.2~s.
Together, these results suggest that explicitly incorporating subject-specific cortical geometry enables CANDLE to generalize to unseen anatomies while preserving both the spatial localization and temporal dynamics of cortical source activity.
Full numerical results and further qualitative results are reported in Table~\ref{tab:raw_sim} and Figure~\ref{fig:add_sim} in Appendix~\ref{appendix: additiol_qua}, respectively.

\subsection{Intracranial stimulation localization}
\label{section: eval_spes}
\paragraph{Experimental setup.}
To evaluate ESI under controlled perturbations with anatomically known source locations, we used the dataset of \citet{Parmigiani2022}, which contains simultaneous scalp EEG and stereoelectroencephalography (SEEG) recordings during single-pulse electrical stimulation (SPES).
Specifically, we evaluated whether the intracranial stimulation site could be localized from the corresponding scalp EEG, as illustrated in Figure~\ref{fig:exp_spes}(a).
We included data from 32 patients with drug-resistant focal epilepsy, corresponding to 291 stimulation runs in total (9.1 per subject on average).
Each run comprised an average of 31.5 SPES trials delivered at the same stimulation site.
Following \citet{GBF}, we averaged the EEG responses across trials within each run and used the resulting evoked response as input to each ESI method.

For evaluation, we considered the estimated source activity within a 10-ms window centered on SPES onset. This window excludes subsequent evoked responses (e.g., the N1 and N2 components~\citep{N1_N2}), which reflect propagation to remote regions rather than activity at the stimulation site.
Localization performance was quantified using spatial dispersion and peak distance, which measure the spatial spread of the estimated source activity and the distance between its peak and the known SPES site, respectively.
Both metrics were averaged across runs within each subject.
Detailed definitions are provided in Appendix~\ref{appendix: evaluation metrics}.

\begin{figure}
    \centering
    \includegraphics[width=\linewidth]{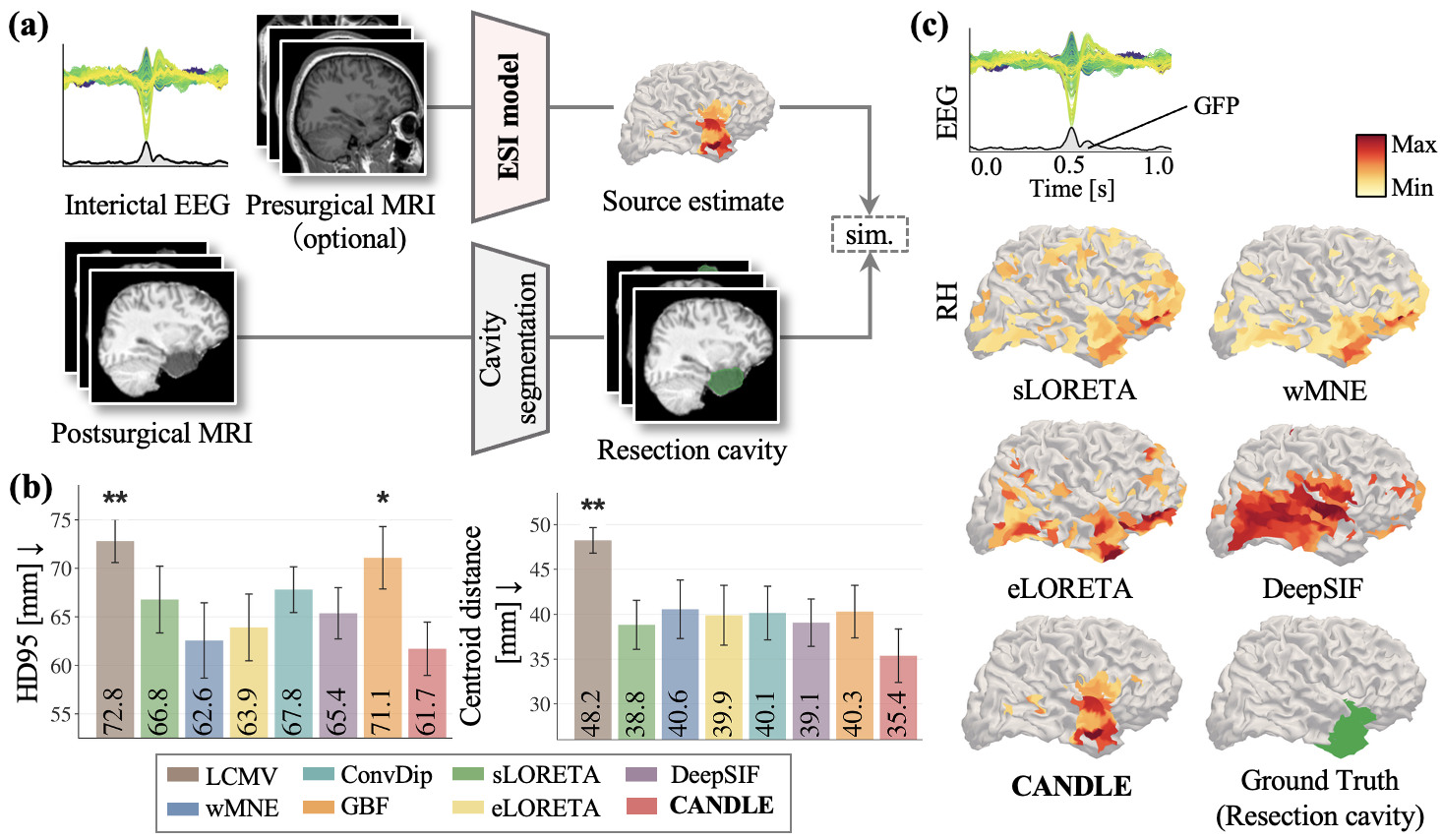}
    \vspace{-1.5em}
    \caption{
    Epileptogenic zone (EZ) estimation.
    (a) Schematic of presurgical EZ localization from interictal EEG.
    (b) Quantitative comparison of ESI methods using HD95 and centroid distance.
    Error bars indicate SEM.
    Statistical significance relative to CANDLE was assessed using two-sided Wilcoxon signed-rank tests (Holm corrected, * $p<0.05$, ** $p<0.01$).
    (c) Qualitative comparison for a representative patient, showing the interictal EEG and estimated source distributions.
    }
    \label{fig:exp_epilesy}
    \vspace{-1em}
\end{figure}

\paragraph{Results.}
Figure~\ref{fig:exp_spes}(b) summarizes the quantitative results for intracranial stimulation localization.
We evaluated spatial dispersion and peak distance across all stimulation runs and further stratified peak distance by SPES site depth.
Across all runs, CANDLE achieved a spatial dispersion of 54.1~mm and a peak distance of 32.9~mm, the lowest errors among all methods and improvements of 2.5~mm and 3.6~mm over the second-best baseline, respectively.
These improvements were statistically significant over five of the seven baselines ($p<0.01$); the differences from sLORETA and GBF were not significant.
When stratified by SPES-site depth, CANDLE achieved a peak distance of 32.48~mm for shallow stimulation sites at depths of 0--10~mm, achieving the second-best performance behind GBF (29.20~mm).
Notably, the localization errors of the baseline methods generally increased as the stimulation site became deeper, whereas CANDLE remained comparatively robust to source depth and achieved the best performance in all depth ranges beyond 10~mm.

Figure~\ref{fig:exp_spes}(c) presents qualitative results for two representative stimulation runs, showing the EEG signals and the estimated SPES locations on glass-brain maps.
In both cases, the ground-truth SPES sites (black stars) were located in the right temporal lobe, and the CANDLE estimates (red circles) were closest to the stimulation sites among all methods.
% Additional qualitative results are shown in Appendix~\ref{appendix: additional_spes_qual}.
Together, these results suggest that the source activity prior learned by CANDLE exclusively from simulated data transfers to empirical EEG recordings.
In particular, its robustness to SPES site depth suggests that the learned null-space prior can complement limited scalp information for deeper sources, enabling plausible localization under challenging source configurations.
Complete quantitative results are listed in Table~\ref{tab:raw_spes} in Appendix~\ref{appendix: additiol_qua}, together with further representative runs in Figure~\ref{fig:add_spes}.

\subsection{Epileptogenic zone estimation}
% \vspace{-0.5em}
\label{section: epilepsy}

\paragraph{Experimental setup.}
To evaluate ESI in a clinically relevant setting, we considered epileptogenic zone (EZ) estimation from presurgical interictal EEG signals.
The EZ is defined as the minimum cortical tissue whose resection or disconnection is necessary to achieve seizure freedom~\citep{zijlmans2019changing}.
Although the EZ cannot be directly observed preoperatively, ESI of interictal epileptiform discharges (IEDs) observed in interictal EEG provides complementary information for estimating epileptogenic regions and is widely used in presurgical planning~\citep{pDeepSIF, DeepSIF}.

Accordingly, we evaluated CANDLE using the dataset of \citet{vorderwulbecke2025}, which provides presurgical EEG containing IEDs together with postsurgical T1-weighted MRI and surgical outcomes.
% Specifically, we evaluated whether the epileptogenic region could be localized from presurgical EEG, as illustrated in Figure~\ref{fig:exp_epilesy}(a).
Specifically, we evaluated whether the epileptogenic region could be localized from presurgical EEG together with the presurgical MRI for methods requiring subject-specific anatomy, as illustrated in Figure~\ref{fig:exp_epilesy}(a).
We included 22 patients with favorable 12-month postsurgical outcomes, defined as International League Against Epilepsy (ILAE)~\citep{ILAE} classes 1 or 2.
Because a favorable surgical outcome suggests that the clinically relevant EZ was encompassed by the resection, we used the postsurgical resection cavity as a reference for the EZ.
The resection cavity was segmented from each postsurgical T1-weighted MRI using MELD-PostOp~\citep{MELD-PostOp}, a U-Net-based model for postoperative resection-cavity segmentation.

For evaluation, we obtained the source region by temporal pooling and Otsu's thresholding~\citep{Otsu}, following Subsection~\ref{section: eval_sim}.
We evaluated localization using HD95 and centroid distance rather than overlap-based metrics (e.g., IoU), because surgical resections may extend beyond the true EZ to ensure complete removal of epileptogenic tissue.
Detailed definitions are provided in Appendix~\ref{appendix: evaluation metrics}.

\begin{figure}
    \centering
    \includegraphics[width=\linewidth]{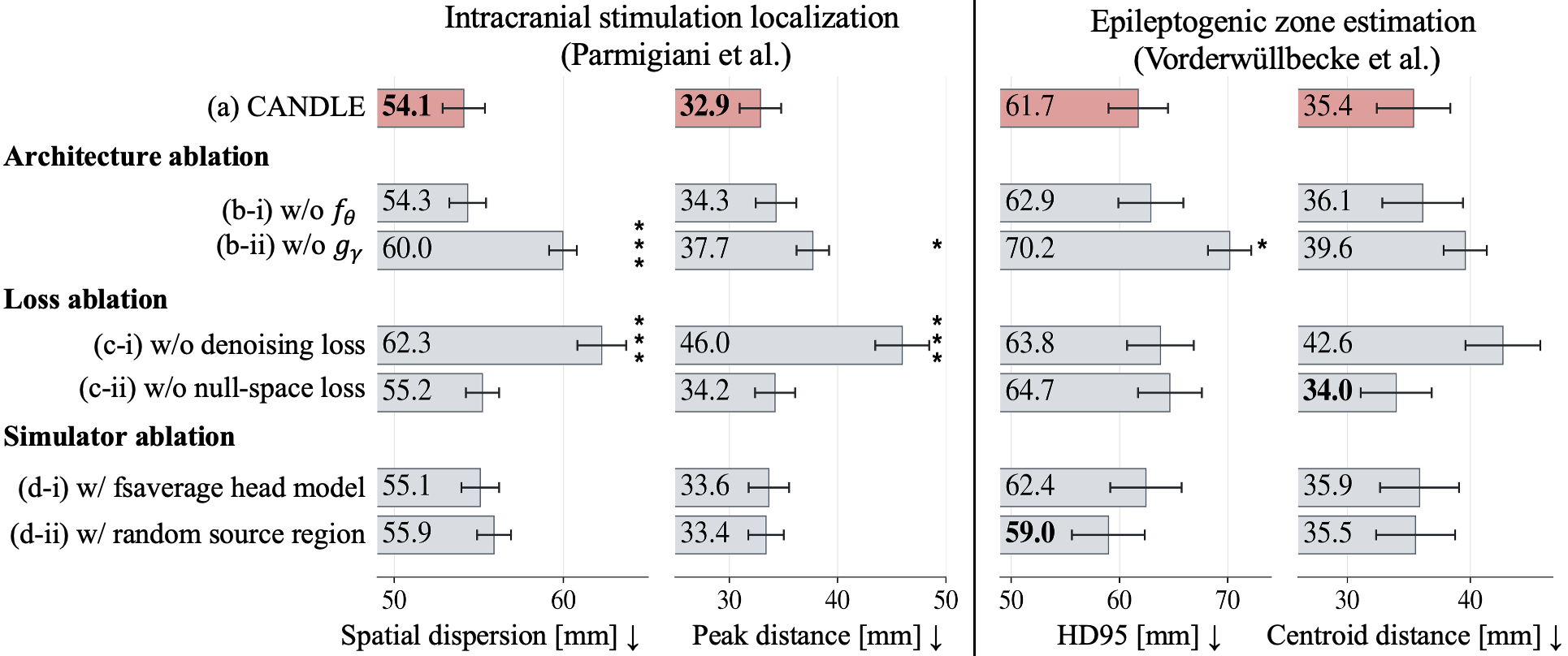}
    \vspace{-1em}
    \caption{
    Ablation studies of CANDLE. Performance of the full model (a) and variants ablating the architecture (b-i, b-ii), training objective (c-i, c-ii), and simulator design (d-i, d-ii), evaluated on intracranial stimulation localization (spatial dispersion and peak distance) and epileptogenic zone estimation (HD95 and centroid distance). Error bars indicate SEM.
    }
    \vspace{-1em}
    \label{fig: ablation}
\end{figure}

\paragraph{Results.}
Figure~\ref{fig:exp_epilesy}(b) presents the quantitative results for EZ estimation.
CANDLE achieved the lowest errors among all methods for both HD95 and centroid distance.
Specifically, CANDLE obtained an HD95 of 61.7~mm and a centroid distance of 35.4~mm, improving the second-best baseline by 0.9~mm and 3.4~mm, respectively.
Compared with LCMV, CANDLE achieved significant improvements in both metrics ($p<0.01$), and significantly improved HD95 over GBF ($p<0.05$).

Figure~\ref{fig:exp_epilesy}(c) further presents qualitative results for a representative patient whose resection involved the anterior temporal lobe.
Given the corresponding presurgical interictal EEG, sLORETA, wMNE, and eLORETA produced spatially diffuse estimates extending beyond the resected region, while DeepSIF localized activity predominantly outside the resection.
In contrast, although the CANDLE estimate showed some posterior extension along the temporal lobe, its dominant source region was largely concentrated around the resected anterior temporal region.
% Additional qualitative results are provided in Appendix~\ref{appendix: additional_epilepsy_qual}.
Together, these results extend the zero-shot transfer observed under controlled stimulation in Section~\ref{section: eval_spes} to a more clinically realistic setting.
Appendix~\ref{appendix: additiol_qua} reports the corresponding full numerical results (Table~\ref{tab:raw_epilepsy}) and qualitative examples for additional patients (Figure~\ref{fig:add_epilepsy}).

\subsection{Ablation Studies}
\vspace{-0.5em}
Figure~\ref{fig: ablation} summarizes three ablation studies evaluated on intracranial stimulation localization~\citep{Parmigiani2022} and epileptogenic zone estimation~\citep{vorderwulbecke2025}.
We use spatial dispersion and peak distance for the former, and HD95 and centroid distance for the latter.

\paragraph{Architecture ablation.}
To validate the architecture of CANDLE, we compared (a) the full model with variants removing (b-i) the denoising model $f_\theta$ or (b-ii) the null-space estimator $g_\gamma$.
Both ablations degraded performance across four metrics, with peak distance increasing by 1.4 and 4.8~mm for Models (b-i) and (b-ii), respectively.
The degradation was consistently larger for Model (b-ii), with statistically significant increases relative to the full model in spatial dispersion ($p<0.001$), peak distance ($p<0.05$), and HD95 ($p<0.05$).
These results suggest that both $f_\theta$ and $g_\gamma$ contribute to CANDLE's performance, while the larger impact of removing $g_\gamma$ highlights the importance of estimating source activity in the null space that is unobservable from scalp measurements.

\paragraph{Loss ablation.}
We next examined the training objective by comparing (a) the full model with variants excluding either (c-i) the denoising loss or (c-ii) the null-space loss in Equation~(\ref{equation: loss}).
Model (c-i) degraded all four metrics, including a 13.1-mm increase in peak distance. The degradation in spatial dispersion and peak distance was significant ($p<0.001$).
By contrast, Model (c-ii) degraded three of the four metrics, except centroid distance, with peak distance increasing by 1.3~mm.
These results suggest that both losses contribute to CANDLE's optimization, with the denoising loss playing a more critical role.
This is likely because $f_\theta$ precedes the range--null space decomposition, such that inadequate denoising undermines the effectiveness of the entire subsequent source reconstruction.

\paragraph{Simulator ablation.}
Finally, we assessed the simulator design using two variants of the full simulator in Section~\ref{section: Simulator_method}: (d-i) a fixed fsaverage head model~\citep{fsaverage} and (d-ii) random source regions of interest (ROIs).
Model (d-i) degraded performance across all four metrics, whereas Model (d-ii) degraded three of the four metrics, with HD95 as the exception.
In particular, peak distance increased by 0.7 and 0.5~mm for Models (d-i) and (d-ii), respectively.
These results suggest that both anatomical diversity across subjects and neurobiologically grounded source regions contribute to effective training of CANDLE.

%% file: 07_Discussion_and_Conclusion.tex
\section{Conclusion and Limitations}
\vspace{-0.5em}
\paragraph{Conclusion.}
In this study, we addressed electrophysiological source imaging (ESI).
We proposed CANDLE, a learning-based ESI model that accounts for subject-specific cortical geometry by learning a prior over the unobservable null-space component of source activity.
To train CANDLE, we further developed a whole-brain simulator that generates neurobiologically grounded source activity across more than 1,100 subject-specific cortical geometries. 
Without further tuning, CANDLE outperformed prior ESI methods on simulated source estimation and two empirical tasks: intracranial stimulation localization and epileptogenic zone estimation.

\paragraph{Limitations.}
CANDLE has several limitations that motivate future work.
(i) The current implementation considers EEG recorded with a fixed montage~\citep{BioSemi64} and represents cortical source activity using the Hagmann/Lausanne parcellation~\citep{hagmann2008mapping}. Extending the framework to accommodate diverse sensor configurations and source-space parcellations would broaden its applicability to recording settings and potentially other electrophysiological modalities, such as MEG.
(ii) Our whole-brain simulator models cortical dynamics using a single canonical neural mass model, the Jansen–Rit model, although different neural dynamics models may better capture different physiological regimes. Future work could incorporate diverse neural dynamics models to learn more generalizable source priors and potentially improve sim-to-real transfer.

%% file: 08_mandatory_section.tex
\subsection*{AI use statement}

In this work, we used generative AI tools to help develop theoretical models or conceptual frameworks, design or provide feedback on research methodology or experiments, and implement methods. Additionally, we used generative AI tools for sourcing and searching for information, identifying relevant literature, and editing the paper to improve readability. 
We did not use generative AI tools to generate synthetic datasets, formulate mathematical claims, provide critical ingredients for proving mathematical claims, assist in writing proofs, propose or refine hypotheses, assist with translation, clean or reformat datasets, support qualitative or thematic data analysis, or interpret results. 
AI-assisted implementations were reviewed and tested by the authors, and AI-assisted literature retrieval was independently verified against the original sources. 
We take responsibility for the final content of this work, including text, claims, or artifacts produced with the aid of generative AI.

\subsection*{Ethics statement}

This study involved only secondary analyses of previously collected human neuroimaging and electrophysiological datasets.
We used data from the Human Connectome Project~\citep{HCP} and the datasets of \citet{Parmigiani2022} and \citet{vorderwulbecke2025}.
No new data from human participants were collected as part of this work.
All data were used in accordance with the applicable data-use terms and the ethical approvals reported in the original studies.

\subsection*{Reproducibility statement}

We provide the methodological details required to reproduce CANDLE and the whole-brain simulator in Section~\ref{section: CANDLE_method} and Appendix~\ref{appendix: Implenentation}, including the model architecture, training configuration, data preprocessing, head-model construction, and neural mass model parameters.
The experimental protocols and evaluation metrics are described in Section~\ref{section: experiments} and Appendix~\ref{appendix: evaluation metrics}, respectively.
The supplementary materials additionally include the implementation code and trained model checkpoints used in our experiments.

%% file: appendix.tex
\section{Related Works}
\paragraph{Electrophysiological source imaging.}
Electrophysiological source imaging (ESI) methodologies and their applications have been systematically reviewed by \citet{ESI_review2} and \citet{ESI_review}. 
Because ESI is severely ill-posed, conventional approaches address this non-uniqueness by imposing explicit spatial constraints on the cortical source space. 
Early methods such as MUSIC \citep{MUSIC} restrict source activity to one or a few equivalent current dipoles, whereas minimum-norm–based estimators such as sLORETA \citep{sLORETA} and dSPM \citep{dSPM} impose depth-dependent weighting on distributed source estimates.
More recent methods introduce richer structured priors over the source space.
FAST-IRES~\citep{FAST-IRES} promotes source sparsity under a temporal basis derived from the observed EEG signals, yielding spatially localized source estimates.
GBF~\citep{GBF} regresses the weights of subject-specific eigenmodes derived from cortical geometry to obtain spatially smooth source estimates.

In contrast to these explicitly constrained formulations, learning-based ESI models seek to acquire data-driven priors over source activity, thereby reducing reliance on manually specified spatial constraints.
ConvDip \citep{ConvDip}, an early representative, learns a CNN-based mapping from EEG topographies to source activity. 
DeepSIF~\citep{DeepSIF} further demonstrates that a model trained exclusively on synthetic data generated using biophysical computational models can transfer to real recordings.
Subsequent work has extended this paradigm in several directions, with XDL-ESI~\citep{XDL-ESI} introducing explainability into learning-based ESI, while MMDF-ANN~\citep{MMDF-ANN} improves source-estimation accuracy by jointly exploiting multimodal EEG/MEG signals.
Most existing learning-based methods, however, are trained in a common template source space such as fsaverage~\citep{fsaverage}, and their application to individual subjects may therefore require post-training adaptation to subject-specific anatomy~\citep{pDeepSIF}.

\paragraph{Brain activity simulators.}
Mathematical models of brain activity span multiple spatial scales, from biophysically detailed single-neuron models such as Hodgkin–Huxley \citep{HodgkinHuxley} to macroscopic neural field models of large-scale cortical dynamics \citep{NeuralFieldModel}. Between these scales, neural mass models (NMMs) describe the aggregate dynamics of interacting neuronal populations and are well suited to electrophysiological activity at the spatial scale relevant to ESI. Representative NMMs include the Wilson–Cowan \citep{wilson1972excitatory} and Jansen–Rit \citep{Jansen-Rit} models, with subsequent variants developed for specific regimes such as epileptiform activity \citep{wendling2002epileptic} and sleep-related dynamics \citep{coronel2026multi}.

More recently, brain activity simulations have been extended to the whole-brain scale by embedding NMMs at the nodes of structural connectomes derived from diffusion MRI \citep{lynn2019physics}. Frameworks such as BrainPy \citep{wang2023brainpy}, The Virtual Brain \citep{TVB}, and Virtual Brain Inference \citep{VBI} facilitate such simulations. However, these simulators are primarily designed to model brain dynamics and often abstract away anatomical variability and the spatial organization of cortical activity, limiting their suitability for applications requiring anatomically realistic simulations across subjects, such as training learning-based ESI models.

\section{Detailed Procedure of CANDLE}

\subsection{Lead-field matrix calculation}
\label{appendix: Leadfied calc}

Among various head-modeling approaches for constructing the lead-field matrix $\bm{L}$, CANDLE derives $\bm{L}$ from the subject-specific anatomical image $\mathcal{G}$ by modeling volume conduction through the head tissues using a boundary element model (BEM)~\citep{hallez2007review}.
Specifically, we first segment the tissue boundaries relevant to EEG generation and volume conduction from $\mathcal{G}$ using a watershed-based segmentation procedure. These surfaces include the white-matter surface, which defines the cortical source space, as well as the inner-skull, outer-skull, and scalp surfaces through which the electrical potentials propagate before reaching the EEG electrodes. Each tissue boundary is subsequently tessellated to construct a surface-based head model, from which volume conduction is modeled using a three-layer BEM.
We assign conductivities of $0.3$, $0.006$, and $0.3~\mathrm{S/m}$ to the brain, skull, and scalp compartments, respectively.

BEM formulations can incorporate different numbers and configurations of tissue compartments. In this work, we adopt a three-layer model. \citet{birot2014head} showed that increasing the complexity of volume-conduction models has only a negligible effect on source localization performance for ESI with empirical EEG recordings. Based on this finding, we employ a three-layer BEM to reduce computational cost while preserving the anatomical detail relevant to EEG propagation.

We further discretize the cortical source space according to the Hagmann/Lausanne parcellation~\citep{hagmann2008mapping}. Previous physiological studies have suggested that cortical activity must be synchronized over an area of approximately $6$--$10~\mathrm{cm}^{2}$ to generate scalp potentials of sufficient magnitude to be detectable by EEG~\citep{tao2005intracranial}. Motivated by this spatial scale, we model the cortical surface using the $N=994$ regions of the Hagmann/Lausanne parcellation, each covering approximately $1.5~\mathrm{cm}^{2}$ of cortical surface. The resulting subject-specific model defines $\bm{L}$, which maps cortical source activity to EEG measurements.

\subsection{Additional Methodological Details}
\label{appendix: Implenentation}

\begin{wrapfigure}{r}{0.4\textwidth}
    \vspace{-2em}
    \centering
    \includegraphics[width=\linewidth]{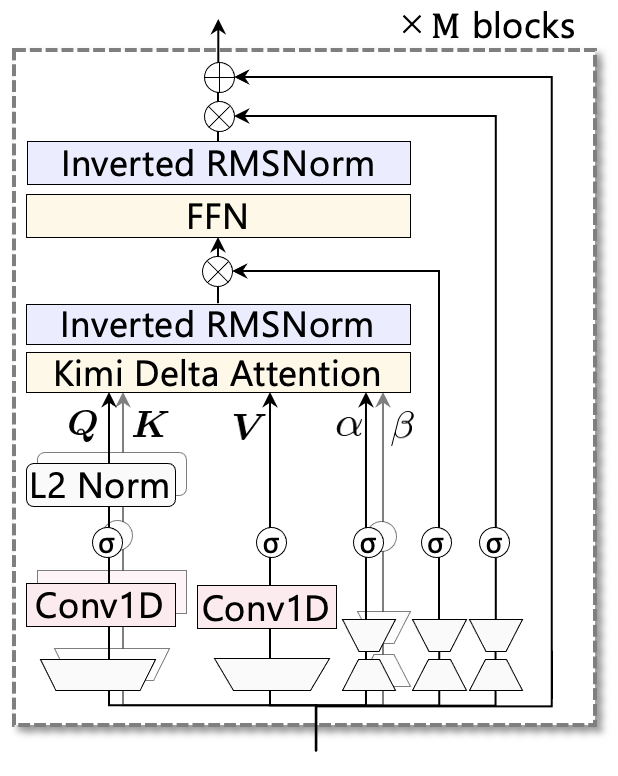}
    \vspace{-2em}
    \caption{Architectures adopted for the denoising model $f_{\theta}$ and null-space estimation model $g_{\gamma}$ in CANDLE.}
    \label{fig:kda_arch}
    \vspace{-1.5em}
\end{wrapfigure}

\paragraph{Model configuration.}
The architectures adopted for the denoising model $f_{\theta}$ and null-space estimation model $g_{\gamma}$ are illustrated in Figure~\ref{fig:kda_arch}.
As the backbone, we employ Kimi Delta Attention~\citep{kimilinear}, a cutting-edge architecture from the family of Delta Rule-based models~\citep{DeltaRule} that have demonstrated strong sequence-modeling performance across diverse modalities.
We extend this backbone with a gated attention unit-based module~\citep{GAU}, which has also shown effectiveness in sequence modeling.
We also adopt inverted normalization along the temporal dimension, following prior works~\citep{RevIn, iTransformer} showing that this strategy improves robustness to non-stationarity in time-series modeling.
We then stack $M$ such blocks to construct the overall architecture.

We trained CANDLE for 20 epochs with a batch size of 64 using AdamW ($\beta_1=0.9,\ \beta_2=0.999$), a learning rate of $1.0\times10^{-4}$, and a weight decay of $1.0\times10^{-2}$. 
We set the sequence length to $T=500$, corresponding to a one-second EEG segment sampled at 500 Hz, and used $M=2$ stacked blocks.
The number of EEG channels was fixed at $C=64$, following the standard BioSemi 64-channel montage~\citep{BioSemi64}, while the cortical source space comprised $N=994$ regions based on the Hagmann--Lausanne parcellation~\citep{hagmann2008mapping}, as detailed in Appendix~\ref{appendix: Leadfied calc}. Under this configuration, CANDLE contained approximately $11.6$~M trainable parameters and required approximately $15.30$~G FLOPs.

We split the simulated dataset into training, validation, and test sets. The training set was used for parameter optimization, the validation set for model selection, and the test set exclusively for final evaluation.
After each training epoch, we evaluated the checkpoint on the validation set using the 95th-percentile Hausdorff distance (HD95), following the procedure described in Subsection~\ref{section: eval_sim} and Appendix~\ref{appendix: evaluation metrics}.
Among the resulting checkpoints, we selected the model achieving the lowest HD95 on the validation set and used this checkpoint for evaluation on the test set.

All experiments were performed on an NVIDIA GeForce RTX 4090 with 24~GB of VRAM and an Intel Core i9-13900KF processor with 64~GB of RAM. 
Training CANDLE under the configuration described above required approximately 1.8~hours.

\paragraph{Data preprocessing.}
EEG signals were downsampled to 500~Hz and band-pass filtered between 0.5 and 45~Hz using a Butterworth filter.
We subsequently applied common average referencing and $z$-score normalization.
For subject-specific anatomical processing, T1-weighted MRIs were processed using FreeSurfer~\citep{freesurfer} to obtain the cortical geometry required for constructing the subject-specific lead-field matrix.

\subsection{Jansen--Rit model}
\label{appendix: NMM}
\input{tables/jansen_rit_config}

We employ the Jansen--Rit neural mass model~\citep{Jansen-Rit} to generate mesoscopic cortical activity for each cortical region in the whole-brain simulator. The model describes the interactions among pyramidal neurons and excitatory and inhibitory interneurons through second-order synaptic dynamics. For cortical region $i$, the observable source activity $y_i(t)$ is defined as the difference between the excitatory and inhibitory postsynaptic potentials,
\begin{equation}
    y_i(t) = y_{1,i}(t) - y_{2,i}(t),
\end{equation}
where $y_{1,i}$ and $y_{2,i}$ denote the excitatory and inhibitory
postsynaptic potentials received by the pyramidal population, respectively. The dynamics are given by
\begin{align}
    \ddot{y}_{0,i}(t)
    &=
    A_i a\,\sigma\!\left(y_i(t)\right)
    -2a\dot{y}_{0,i}(t)-a^2y_{0,i}(t), \\
    \ddot{y}_{1,i}(t)
    &=
    A_i a
    \left[
        C_2\sigma\!\left(C_1y_{0,i}(t)\right)
        + H_i(t)
        + \xi_i(t)
    \right]
    -2a\dot{y}_{1,i}(t)-a^2y_{1,i}(t), \\
    \ddot{y}_{2,i}(t)
    &=
    Bb
    \left[
        C_4\sigma\!\left(C_3y_{0,i}(t)\right)
    \right]
    -2b\dot{y}_{2,i}(t)-b^2y_{2,i}(t),
\end{align}
where $y_{0,i}$ denotes the excitatory postsynaptic potential generated by the pyramidal population, $H_i(t)$ denotes the interregional excitatory input from other cortical regions, $\xi_i(t)$ denotes the stochastic input, and $\sigma(\cdot)$ converts the average membrane potential into the population firing rate:
\begin{equation}
    \sigma(v)
    =
    \frac{e_{\max}}
    {1+\exp\{r(v_0-v)\}}.
\end{equation}
The remaining model parameters are summarized in Table~\ref{tab:jansen_rit_parameters}.

To extend the local Jansen--Rit dynamics to the whole cortex, we instantiate one neural mass model at each of the $N$ cortical regions and couple their pyramidal populations according to the structural connectivity of the Hagmann--Lausanne parcellation~\citep{hagmann2008mapping}. The interregional input $H_i(t)$ is computed as
\begin{equation}
H_i(t)
=
\sum_{j=1}^{N}
W_{ij}
\sigma\left(y_j(t)\right),
\end{equation}
where $W_{ij}$ denotes the normalized structural connectivity from region $j$ to region $i$, with $W_{ii}=0$.
The resulting regional activities are concatenated to form the whole-brain cortical source activity
$\bm{Y}(t)=[y_1(t),\ldots,y_N(t)]^\top$.

Following ~\citet{DeepSIF}, the source configurations defined in Subsection~\ref{section: Simulator_method} are implemented by locally modulating the excitatory synaptic gain $A_i$. For a sampled source configuration $\mathcal{S}$, we set
\begin{equation}
    A_i =
    \begin{cases}
        3.50~\mathrm{mV}, & i\in\mathcal{S},\\
        3.25~\mathrm{mV}, & i\notin\mathcal{S},
    \end{cases}
\end{equation}
while keeping all other Jansen--Rit parameters unchanged. Thus, source
activity is generated by selectively increasing the excitatory
postsynaptic response of neurobiologically defined cortical regions rather
than directly injecting an artificial waveform into the source space.

The stochastic differential equations were numerically integrated using the stochastic Heun method at a sampling rate of $2{,}000$~Hz. Each simulation began with a $1$-s burn-in period to reduce dependence on the initial state. This period was discarded, and the subsequent $10$~s of simulated activity were used for analysis. We generated three independent trials for each source configuration under each of three noise conditions, resulting in $30$~s of simulated activity per noise condition. The simulated cortical activity was subsequently downsampled from $2{,}000$ to $500$~Hz.

% \begin{align}
%     x_i(t)
%     &= x_{1,i}(t) - x_{2,i}(t) \\
%     x_{0,i}(t)
%     &= h_{\rm{e}}(t) * \left((x_{1,i}(t) - x_{2,i}(t)\right)\\
%     x_{1,i}(t)
%     &= h_{\rm{e}}(t) *\left[C_2\sigma(C_1 x_{0,i}(t)) + H_i + \epsilon\right]\\
%     x_{2,i}(t)
%     &= h_{\rm{i}}(t) * \left[C_4\sigma(C_3 x_{0,i}(t))\right]\\
%     h_{\rm{e}}(t)&= Aate^{-at}\\
%     h_{\rm{i}}(t)&=Bbte^{-bt}\\
%     \sigma(v)
%     &= \frac{v_{\textrm{max}}}
%     {1+\textrm{exp}\{r(v_0-v)\}}\\
%     H_i &= \sum_{j}W_{i,j}\sigma\left(x_{1,i}(t) - x_{2,i}(t)\right)
% \end{align}

% \begin{align}
%     x_i(t)
%     &= x_{1,i}(t) - x_{2,i}(t) \\
%     \frac{\partial^2 x_{0,i}(t)}{\partial t}
%     &= Aa\sigma\left(x_{1,i}(t) - x_{2,i}(t)\right)
%     - 2a\frac{\partial x_{0,i}(t)}{\partial t}
%     - a^2x_{0,i}(t) \\
%     \frac{\partial^2 x_{0,i}(t)}{\partial t}
%     &= Aa\left[C_2\sigma(C_1 x_{0,i}(t)) + H_i + \epsilon\right]
%     - 2a\frac{\partial x_{1,i}(t)}{\partial t}
%     - a^2x_{1,i}(t) \\
%     \frac{\partial^2 x_{2,i}(t)}{\partial t}
%     &= Bb\left[C_4\sigma(C_3 x_{0,i}(t))\right]
%     - 2b\frac{\partial x_{2,i}(t)}{\partial t}
%     - b^2x_{2,i}(t) \\
%     \sigma(v)
%     &= \frac{v_{\textrm{max}}}
%     {1+\textrm{exp}\{r(v_0-v)\}
%     }
% \end{align}

\section{Evaluation Metrics}
\label{appendix: evaluation metrics}

We provide detailed definitions of the evaluation metrics used in Section~\ref{section: experiments}.
All spatial distances are computed in the subject-specific anatomical space.

\paragraph{95th-percentile Hausdorff distance (HD95).}
Let $\mathcal{S}$ and $\widehat{\mathcal{S}}$ denote the ground-truth and estimated source regions, respectively.
For two sets of spatial points $\mathcal{A}$ and $\mathcal{B}$, the directed nearest-neighbor distances are defined as
\begin{equation}
    D(\mathcal{A},\mathcal{B})
    =
    \left\{
    \min_{\bm{b}\in\mathcal{B}}
    \|\bm{a}-\bm{b}\|_2
    \,\middle|\,
    \bm{a}\in\mathcal{A}
    \right\}.
\end{equation}
The HD95 is then defined as follows, where $Q_{0.95}(\cdot)$ denotes the 95th percentile:
\begin{equation}
    \mathrm{HD95}(\mathcal{S},\widehat{\mathcal{S}})
    =
    \max
    \left\{
    Q_{0.95}\!\left(D(\mathcal{S},\widehat{\mathcal{S}})\right),
    Q_{0.95}\!\left(D(\widehat{\mathcal{S}},\mathcal{S})\right)
    \right\}.
\end{equation}

\paragraph{Dice score.}
The Dice score quantifies the spatial overlap between the estimated and ground-truth source regions and is defined as
\begin{equation}
    \mathrm{Dice}(\mathcal{S},\widehat{\mathcal{S}})
    =
    \frac{2|\mathcal{S}\cap\widehat{\mathcal{S}}|}
    {|\mathcal{S}|+|\widehat{\mathcal{S}}|}.
\end{equation}

\paragraph{Spatial dispersion.}
For intracranial stimulation localization, we quantify the spatial spread of the estimated source activity relative to the stimulation site using spatial dispersion (SD).
Let $\bm{r}_{\mathrm{stim}}$ denote the stimulation location, $\bm{r}_k$ the location of the $k$-th cortical source, and
$d_k=\|\bm{r}_k-\bm{r}_{\mathrm{stim}}\|_2$.
Following the definition of \citet{vorderwulbecke2025}, spatial dispersion is calculated as follows, where $R_k$ denotes the estimated activity of the $k$-th source at the time point of maximum source activity:
\begin{equation}
    \mathrm{SD}
    =
    \sqrt{
    \frac{
    \sum_{k=1}^{N} d_k^2 R_k^2
    }{
    \sum_{k=1}^{N} R_k^2
    }}.
\end{equation}

\paragraph{F1 score across IoU thresholds.}
IoU measures the overlap between the estimated and reference source regions relative to their union.
For each IoU threshold from 0.1 to 0.9 in increments of 0.1, an estimate was considered correct if its IoU with the ground-truth region exceeded the threshold, and the corresponding F1 score was computed.

\paragraph{Peak distance.}
Peak distance is the Euclidean distance between the known stimulation site and the cortical source with the maximum estimated activity.

\paragraph{Centroid distance.}
Centroid distance is the Euclidean distance between the spatial centroids of the estimated and ground-truth source regions.

\section{Additional experimental results}
\label{appendix: additiol_qua}

\begin{figure}
    \centering
    \includegraphics[width=\linewidth]{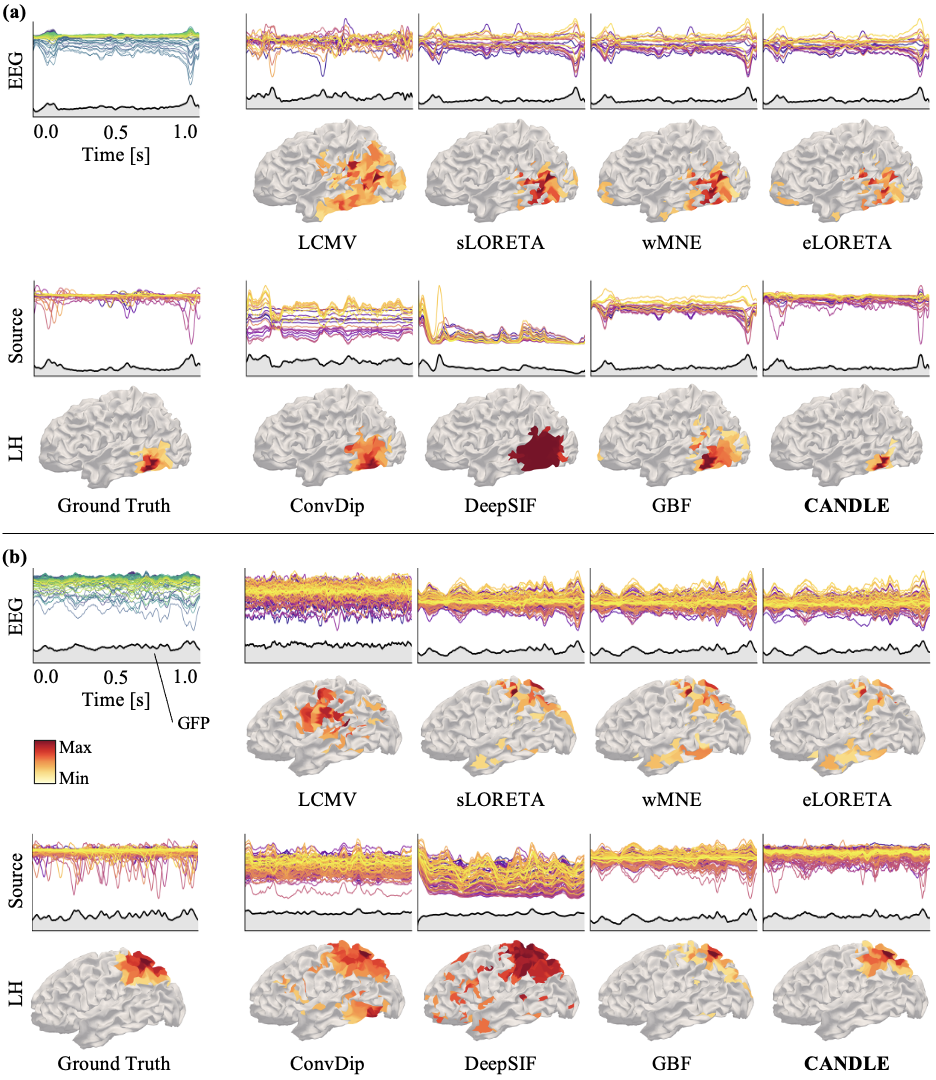}
    \caption{
    Additional qualitative results for simulated source activity estimation. Two representative test samples, (a) and (b). In each panel, the leftmost column shows the EEG input with its global field power (GFP), the ground-truth source activity, and its temporally pooled spatial map. The remaining columns show, for each of the eight ESI methods, the estimated source activity with its GFP and the corresponding temporally pooled spatial map.
    }
    \label{fig:add_sim}
\end{figure}

\begin{figure}
    \centering
    \includegraphics[width=\linewidth]{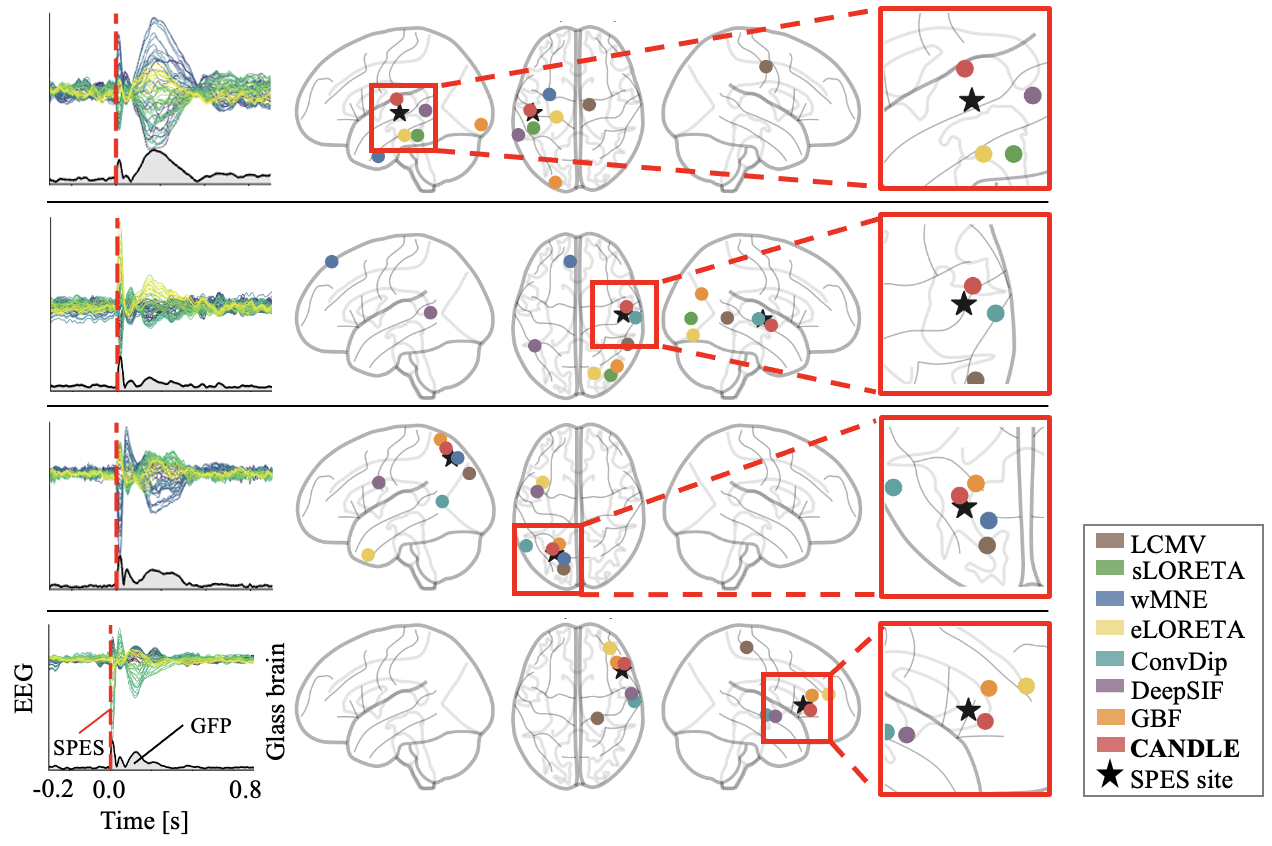}
    \caption{
    Additional qualitative results for intracranial stimulation localization. 
    Four representative stimulation runs from different patients. 
    In each row, the left panel shows the input EEG with its GFP, where the dashed line marks SPES onset. The remaining panels show the estimated SPES locations on glass-brain maps in left lateral, superior, and right lateral views.
    }
    \label{fig:add_spes}
\end{figure}

\begin{figure}
    \centering
    \includegraphics[width=\linewidth]{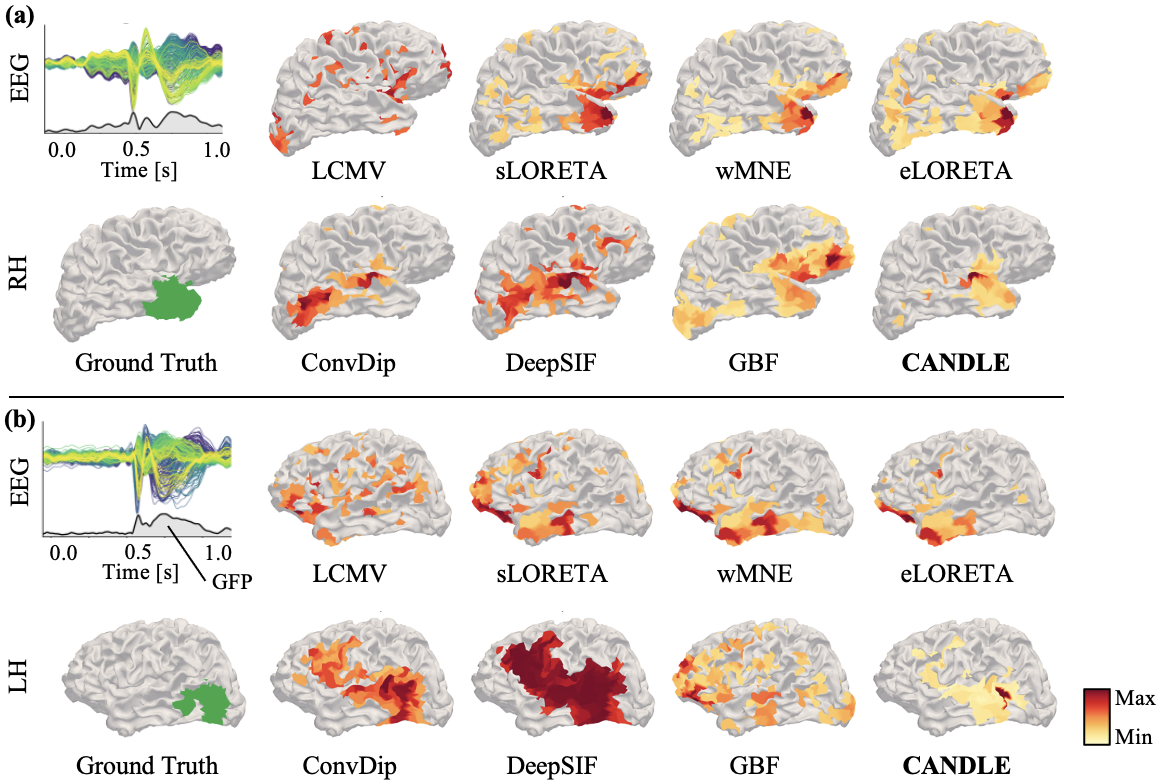}
    \caption{
    Additional qualitative results for epileptogenic zone estimation. Two representative patients with (a) a right-hemisphere (RH) and (b) a left-hemisphere (LH) resection. For each patient, the interictal EEG with its GFP is shown together with the temporally pooled source estimates of all eight ESI methods and the postsurgical resection cavity (green) used as the reference EZ.
    }
    \label{fig:add_epilepsy}
\end{figure}

\paragraph{Additional qualitative results.}
Figures~\ref{fig:add_sim}, \ref{fig:add_spes}, and \ref{fig:add_epilepsy} present additional qualitative results for simulated source activity estimation, intracranial stimulation localization, and epileptogenic zone estimation, respectively.
Figure~\ref{fig:add_sim} shows two additional test samples, comparing all eight ESI methods.
As in Figure~\ref{fig:exp_sim}(b), the conventional methods and the learning-based baselines produced spatially diffuse or displaced estimates, whereas CANDLE recovered a source region whose location and extent closely matched the ground truth.
Figure 9 shows four additional stimulation runs from different patients, with the estimated SPES locations displayed in left lateral, superior, and right lateral views.
Figure~\ref{fig:add_epilepsy} shows two additional patients with right- and left-hemisphere resections. 
In both cases, the dominant source region estimated by CANDLE lay within or adjacent to the resection cavity.

\paragraph{Full quantitative results.}  
Tables~\ref{tab:raw_sim}, \ref{tab:raw_spes}, and \ref{tab:raw_epilepsy} report the complete numerical results underlying Figures~\ref{fig:exp_sim}(a), \ref{fig:exp_spes}(b), and \ref{fig:exp_epilesy}(b), respectively.
Across all three tables, CANDLE achieved the best or second-best value in every metric, and the only case in which a baseline outperformed CANDLE was the shallowest SPES sites (0--10~mm), where GBF yield a lower peak distance.

\input{tables/eval_sim_quant}
\input{tables/eval_spes}
\input{tables/eval_epilepsy}

%% file: tables/jansen_rit_config.tex
\begin{table}[t]
    \centering
    \caption{Parameters of the Jansen--Rit model.}
    \label{tab:jansen_rit_parameters}
    \begin{tabularx}{\linewidth}{ l l X}
        \toprule
        Symbol & Value & Description \\
        \midrule
        $A_i$ 
        & $3.25 ~\mathrm{mV}$ 
        & Average amplitude of the excitatory postsynaptic potential\\
        $B$ 
        & $22~\mathrm{mV}$ 
        & Average amplitude of the inhibitory postsynaptic potential\\
        $a$ 
        & $100~\mathrm{s}^{-1}$ 
        & Time constant of the excitatory synaptic response \\
        $b$ 
        & $50~\mathrm{s}^{-1}$ 
        & Time constant of the inhibitory synaptic response \\
        $C_1, C_2$ 
        & $135,\ 108$ 
        & Average pyramidal cell--excitatory interneuron synaptic connectivity\\
        $C_3, C_4$ 
        & $33.75,\ 33.75$ 
        & Average pyramidal cell--inhibitory interneuron synaptic connectivity\\
        $e_{\max}$ 
        & $5~\mathrm{Hz}$ 
        & Maximum firing rate \\
        $v_0$ 
        & $6~\mathrm{mV}$ 
        & Membrane potential at half-maximum firing rate \\

        $r$ 
        & $0.56~\mathrm{mV}^{-1}$ 
        & Slope of the sigmoid function at $v_0$ \\
        \bottomrule
    \end{tabularx}
\end{table}

%% file: tables/eval_sim_quant.tex
\begin{table*}[t]
\centering
\caption{Full quantitative results for simulated source activity estimation.
\textbf{Bold} and \underline{underlined} values
indicate the best and second-best performances, respectively.
Statistical significance relative to CANDLE was assessed using two-sided Wilcoxon signed-rank tests (Holm corrected, $\dagger: p < 0.05$, $\ddagger: p < 0.01$, $\dagger\dagger: p < 0.001$). Corresponds to Figure~\ref{fig:exp_sim}(a).}
\label{tab:raw_sim}
\small
\setlength{\tabcolsep}{4pt}
\renewcommand{\arraystretch}{1.2}
\resizebox{\textwidth}{!}{%
\begin{tabular}{lccccccc>{\columncolor{candlebg}}c}
\toprule
Metric & LCMV & sLORETA & wMNE & eLORETA & ConvDip & DeepSIF & GBF & \textbf{CANDLE} \\
\midrule
HD95 [mm] $\downarrow$ & \ms{57.94\sig{\dagger\dagger}}{9.07} & \ms{66.08\sig{\dagger\dagger}}{8.86} & \ms{70.74\sig{\dagger\dagger}}{8.58} & \ms{69.71\sig{\dagger\dagger}}{8.00} & \ms{\underline{41.05}\sig{\dagger\dagger}}{7.35} & \ms{45.12\sig{\dagger\dagger}}{7.93} & \ms{65.31\sig{\dagger\dagger}}{9.27} & \msb{37.26}{9.60} \\
Dice score$\uparrow$ & \ms{0.22\sig{\dagger\dagger}}{0.06} & \ms{0.24\sig{\dagger\dagger}}{0.05} & \ms{0.21\sig{\dagger\dagger}}{0.05} & \ms{0.19\sig{\dagger\dagger}}{0.05} & \ms{\underline{0.33}\sig{\dagger\dagger}}{0.07} & \ms{\underline{0.33}\sig{\dagger\dagger}}{0.08} & \ms{0.26\sig{\dagger\dagger}}{0.05} & \msb{0.48}{0.07} \\
F1@0.1 $\uparrow$ & \ms{0.48\sig{\dagger\dagger}}{0.15} & \ms{0.56\sig{\dagger\dagger}}{0.16} & \ms{0.52\sig{\dagger\dagger}}{0.15} & \ms{0.47\sig{\dagger\dagger}}{0.15} & \ms{\underline{0.67}\sig{\dagger\dagger}}{0.14} & \ms{0.62\sig{\dagger\dagger}}{0.14} & \ms{0.62\sig{\dagger\dagger}}{0.16} & \msb{0.83}{0.11} \\
F1@0.2 $\uparrow$ & \ms{0.31\sig{\dagger\dagger}}{0.13} & \ms{0.34\sig{\dagger\dagger}}{0.14} & \ms{0.26\sig{\dagger\dagger}}{0.13} & \ms{0.21\sig{\dagger\dagger}}{0.13} & \ms{\underline{0.48}\sig{\dagger\dagger}}{0.15} & \ms{0.47\sig{\dagger\dagger}}{0.16} & \ms{0.37\sig{\dagger\dagger}}{0.15} & \msb{0.71}{0.12} \\
F1@0.3 $\uparrow$ & \ms{0.14\sig{\dagger\dagger}}{0.11} & \ms{0.12\sig{\dagger\dagger}}{0.11} & \ms{0.07\sig{\dagger\dagger}}{0.08} & \ms{0.05\sig{\dagger\dagger}}{0.07} & \ms{0.32\sig{\dagger\dagger}}{0.13} & \ms{\underline{0.33}\sig{\dagger\dagger}}{0.14} & \ms{0.14\sig{\dagger\dagger}}{0.10} & \msb{0.58}{0.15} \\
F1@0.4 $\uparrow$ & \ms{0.03\sig{\dagger\dagger}}{0.05} & \ms{0.02\sig{\dagger\dagger}}{0.04} & \ms{0.01\sig{\dagger\dagger}}{0.03} & \ms{0.01\sig{\dagger\dagger}}{0.02} & \ms{0.19\sig{\dagger\dagger}}{0.10} & \ms{\underline{0.22}\sig{\dagger\dagger}}{0.13} & \ms{0.03\sig{\dagger\dagger}}{0.05} & \msb{0.43}{0.15} \\
F1@0.5 $\uparrow$ & \ms{0.00\sig{\dagger\dagger}}{0.01} & \ms{0.00\sig{\dagger\dagger}}{0.01} & \ms{0.00\sig{\dagger\dagger}}{0.00} & \ms{0.00\sig{\dagger\dagger}}{0.00} & \ms{0.09\sig{\dagger\dagger}}{0.08} & \ms{\underline{0.12}\sig{\dagger\dagger}}{0.09} & \ms{0.00\sig{\dagger\dagger}}{0.02} & \msb{0.27}{0.13} \\
F1@0.6 $\uparrow$ & \ms{0.00\sig{\dagger\dagger}}{0.00} & \ms{0.00\sig{\dagger\dagger}}{0.00} & \ms{0.00\sig{\dagger\dagger}}{0.00} & \ms{0.00\sig{\dagger\dagger}}{0.00} & \ms{0.04\sig{\dagger\dagger}}{0.05} & \ms{\underline{0.06}\sig{\dagger\dagger}}{0.07} & \ms{0.00\sig{\dagger\dagger}}{0.01} & \msb{0.15}{0.10} \\
F1@0.7 $\uparrow$ & \ms{0.00\sig{\dagger\dagger}}{0.00} & \ms{0.00\sig{\dagger\dagger}}{0.00} & \ms{0.00\sig{\dagger\dagger}}{0.00} & \ms{0.00\sig{\dagger\dagger}}{0.00} & \ms{0.01\sig{\dagger\dagger}}{0.03} & \ms{\underline{0.02}\sig{\dagger\dagger}}{0.04} & \ms{0.00\sig{\dagger\dagger}}{0.00} & \msb{0.06}{0.07} \\
F1@0.8 $\uparrow$ & \ms{0.00\sig{\dagger\dagger}}{0.00} & \ms{0.00\sig{\dagger\dagger}}{0.00} & \ms{0.00\sig{\dagger\dagger}}{0.00} & \ms{0.00\sig{\dagger\dagger}}{0.00} & \ms{0.00\sig{\dagger\dagger}}{0.01} & \ms{0.00\sig{\dagger\dagger}}{0.02} & \ms{0.00\sig{\dagger\dagger}}{0.00} & \msb{0.02}{0.04} \\
F1@0.9 $\uparrow$ & \ms{0.00\sig{\ddagger}}{0.00} & \ms{0.00\sig{\ddagger}}{0.00} & \ms{0.00\sig{\ddagger}}{0.00} & \ms{0.00\sig{\ddagger}}{0.00} & \ms{0.00\sig{\ddagger}}{0.00} & \ms{0.00\sig{\ddagger}}{0.00} & \ms{0.00\sig{\ddagger}}{0.00} & \msb{0.01}{0.02} \\
\bottomrule
\end{tabular}
}
\end{table*}

%% file: tables/eval_spes.tex
% Requires: booktabs, xcolor[table], array, graphicx
\definecolor{candlebg}{RGB}{252,228,228}
\providecommand{\ms}[2]{\shortstack{#1\\[1pt]{\scriptsize$\pm$#2}}}
\providecommand{\msb}[2]{\shortstack{\textbf{#1}\\[1pt]{\scriptsize$\pm$#2}}}
\providecommand{\sig}[1]{\ensuremath{^{#1}}}

\begin{table*}[t]
\centering
\caption{Full quantitative results for intracranial stimulation localization on the dataset of \citet{Parmigiani2022}.
Boldface, underlining, and significance markers follow the conventions in Table~\ref{tab:raw_sim}.
Corresponds to Figure~\ref{fig:exp_spes}(b).}
\label{tab:raw_spes}
\small
\setlength{\tabcolsep}{4pt}
\renewcommand{\arraystretch}{1.2}
\resizebox{\textwidth}{!}{%
\begin{tabular}{lccccccc>{\columncolor{candlebg}}c}
\toprule
Metric & LCMV & sLORETA & wMNE & eLORETA & ConvDip & DeepSIF & GBF & \textbf{CANDLE} \\
\midrule
Spatial dispersion [mm] $\downarrow$ & \ms{69.71\sig{\dagger\dagger}}{6.59} & \ms{57.83}{6.52} & \ms{61.38\sig{\dagger\dagger}}{7.04} & \ms{62.62\sig{\dagger\dagger}}{6.72} & \ms{59.05\sig{\dagger\dagger}}{7.11} & \ms{64.95\sig{\dagger\dagger}}{6.43} & \ms{\underline{56.60}}{7.07} & \msb{54.11}{7.09} \\
\midrule
\multicolumn{8}{l}{Peak distance [mm] $\downarrow$}
& \cellcolor{candlebg} \\
\quad Overall & \ms{58.18\sig{\dagger\dagger}}{12.64} & \ms{37.47}{13.66} & \ms{48.11\sig{\dagger\dagger}}{17.90} & \ms{50.09\sig{\dagger\dagger}}{13.44} & \ms{41.12\sig{\ddagger}}{12.28} & \ms{46.02\sig{\dagger\dagger}}{14.39} & \ms{\underline{36.47}}{10.98} & \msb{32.86}{10.83} \\
\quad 0--10\,mm & \ms{55.65\sig{\dagger\dagger}}{16.96} & \ms{32.62}{19.49} & \ms{40.84}{21.89} & \ms{49.40\sig{\dagger}}{22.32} & \ms{38.82}{16.89} & \ms{46.95\sig{\dagger\dagger}}{19.47} & \msb{29.20}{16.83} & \ms{\underline{32.48}}{12.32} \\
\quad 10--20\,mm & \ms{55.52\sig{\dagger\dagger}}{19.21} & \ms{34.83}{15.19} & \ms{49.44\sig{\dagger}}{25.81} & \ms{56.27\sig{\dagger\dagger}}{21.66} & \ms{45.59\sig{\dagger}}{20.76} & \ms{46.10\sig{\dagger}}{19.09} & \ms{\underline{36.39}}{13.20} & \msb{31.35}{15.36} \\
\quad 20--30\,mm & \ms{61.25\sig{\dagger\dagger}}{19.67} & \ms{41.53}{16.05} & \ms{54.68\sig{\dagger}}{26.18} & \ms{44.66}{18.45} & \ms{\underline{40.82}}{18.16} & \ms{44.73}{20.10} & \ms{46.88}{17.40} & \msb{36.04}{11.80} \\
\quad 30--45\,mm & \ms{57.01}{9.64} & \ms{\underline{46.48}}{16.33} & \ms{54.48}{20.64} & \ms{62.44\sig{\dagger}}{19.92} & \ms{49.92}{12.18} & \ms{54.45}{12.76} & \ms{51.05\sig{\ddagger}}{16.28} & \msb{38.04}{14.56} \\
\bottomrule
\end{tabular}
}
\end{table*}

%% file: tables/eval_epilepsy.tex
\begin{table*}[t]
\centering
\caption{Full quantitative results for epileptogenic zone estimation on the dataset of \citep{vorderwulbecke2025}. 
Boldface, underlining, and significance markers follow the conventions in Table~\ref{tab:raw_sim}.
Corresponds to Figure~\ref{fig:exp_epilesy}(b).}
\label{tab:raw_epilepsy}
\small
\setlength{\tabcolsep}{4pt}
\renewcommand{\arraystretch}{1.2}
\resizebox{\textwidth}{!}{%
\begin{tabular}{lccccccc>{\columncolor{candlebg}}c}
\toprule
Metric & LCMV & sLORETA & wMNE & eLORETA & ConvDip & DeepSIF & GBF & \textbf{CANDLE} \\
\midrule
HD95 [mm] $\downarrow$ & \ms{72.82\sig{\ddagger}}{10.43} & \ms{66.79}{16.12} & \ms{\underline{62.57}}{18.20} & \ms{63.93}{16.13} & \ms{67.80}{11.08} & \ms{65.37}{12.33} & \ms{71.08\sig{\dagger}}{15.10} & \msb{61.72}{12.88} \\
Centroid distance [mm] $\downarrow$ & \ms{48.25\sig{\ddagger}}{6.69} & \ms{\underline{38.82}}{12.81} & \ms{40.56}{15.27} & \ms{39.89}{15.61} & \ms{40.14}{13.99} & \ms{39.07}{12.36} & \ms{40.29}{13.67} & \msb{35.38}{13.98} \\
\bottomrule
\end{tabular}
}
\end{table*}